%% file: main.tex
\documentclass[10pt,twocolumn,letterpaper]{article}

\usepackage[11]{cvpr} 
\usepackage{bm}

\usepackage [utf8]{inputenc}

\definecolor{cvprblue}{rgb}{0.21,0.49,0.74}
\usepackage[pagebackref,breaklinks,colorlinks,allcolors=cvprblue]{hyperref}

\def\paperID{2977} 
\def\confName{CVPR}
\def\confYear{2025}

\title{Multi-Level Decoupled Relational Distillation for Heterogeneous Architectures}

\author{Yaoxin Yang\\
\and
Peng Ye\\
\and 
Weihao Lin\\
\and
Kangcong Li\\
\and
Yan Wen\\
\and
Jia Hao\\
\and
Tao Chen\\
}
\author{Yaoxin Yang$^1$ \quad
Peng Ye$^1$ \quad
Weihao Lin$^1$ \quad
Kangcong Li$^1$ \quad
Yan Wen$^1$ \quad
\\
Jia Hao$^1$ \quad
Tao Chen$^1$\footnotemark[2]\\
$^1$School of Information Science and Technology, Fudan University\\
{\tt\small yxyang24@m.fudan.edu.cn, eetchen@fudan.edu.cn}
}
\begin{document}
\maketitle
\renewcommand{\thefootnote}{\fnsymbol{footnote}}
\footnotetext[2]{Corresponding authors.}
\vspace{-5mm}
\input{sec/0_abstract}

\input{sec/1_intro}

\input{sec/2_related_work}

\input{sec/3_methodology}
\input{sec/4_experiments}
\input{sec/5_conclution}





\end{document}

%% file: sec/0_abstract.tex
\begin{abstract}

Heterogeneous distillation is an effective way
to transfer knowledge from cross-architecture teacher models to student models. However, existing heterogeneous distillation methods do not take full advantage of the dark knowledge hidden in the teacher's output, limiting their performance.
To this end, we propose a novel framework named \textbf{M}ulti-\textbf{L}evel \textbf{D}ecoupled \textbf{R}elational \textbf{K}nowledge \textbf{D}istillation (\textbf{MLDR-KD}) to unleash the potential of relational distillation in heterogeneous distillation. Concretely, we first introduce Decoupled Finegrained Relation Alignment (DFRA) in both logit and feature levels to balance the trade-off between distilled dark knowledge and the confidence in the correct category of the  heterogeneous teacher model. 
Then, Multi-Scale Dynamic Fusion (MSDF) module is applied to dynamically fuse the projected logits of multiscale features at different stages in student model, further improving performance of our method in feature level. We verify our method on four architectures (CNNs, Transformers, MLPs and Mambas), two datasets (CIFAR-100 and Tiny-ImageNet). Compared with the best available method, our MLDR-KD improves student model performance with gains of up to 4.86\% on CIFAR-100 and 2.78\% on Tiny-ImageNet datasets respectively, showing robustness and generality in heterogeneous distillation. Code will be released soon.
\end{abstract}

%% file: sec/1_intro.tex
\section{Introduction}
\label{sec:intro}

\begin{figure}[t]
  \centering
  \includegraphics[width=0.9\linewidth]{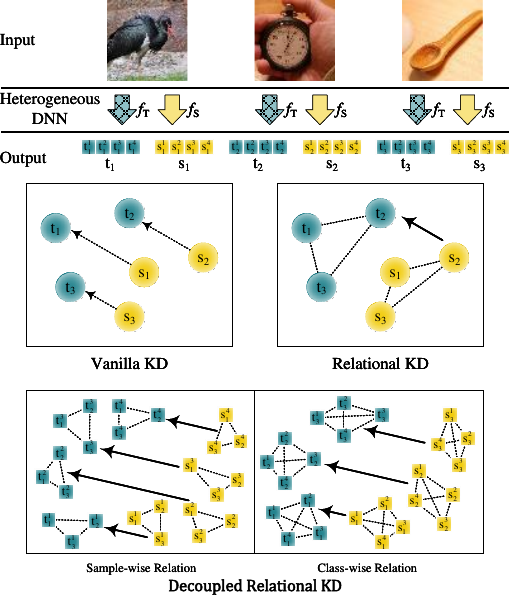}
  \vspace{-2mm}
  \caption{Conceptual comparisons of different knowledge distillation methods.
  Our Decoupled Relational KD first decouples the logits of teacher and student into multiple finegrained relationships between different classes under each sample and different samples under each class, and then aligns the relationships. In our method, Decoupled Relational KD is applied to both logit and multiscale feature levels (namely MLDR-KD).
  }
  \label{fig:header}
  \vspace{-4mm}
\end{figure}

Recently, knowledge distillation (KD)~\cite{hinton2015distilling}, which aims to train a superior lightweight student model by mimicking the teacher model, has been demonstrated to be one of the most effective approaches for model compression~\cite{tang2022patchslimmingefficientvision,zhang2019your}. 
The majority of existing knowledge distillation methods~\cite{tang2022patchslimmingefficientvision,zhang2019your,phuong2019distillation,luan2019msd,zhu2018knowledge} concentrate on the 
distillation between teacher and student models with homogeneous architectures. 
However, this narrow focus limits the widespread use of knowledge distillation. On one hand, there continually emerge new network architectures such as mamba~\cite{gu2023mamba}. On the other hand, there exist various pretrained models that have superior performance but different architectures~\cite{he2022masked,xie2022simmim,beit}.
Consequently, it is essential to explore the potential of knowledge distillation between heterogeneous architectures.

\begin{figure}[t]
  \centering
  \includegraphics[width=0.95\linewidth]{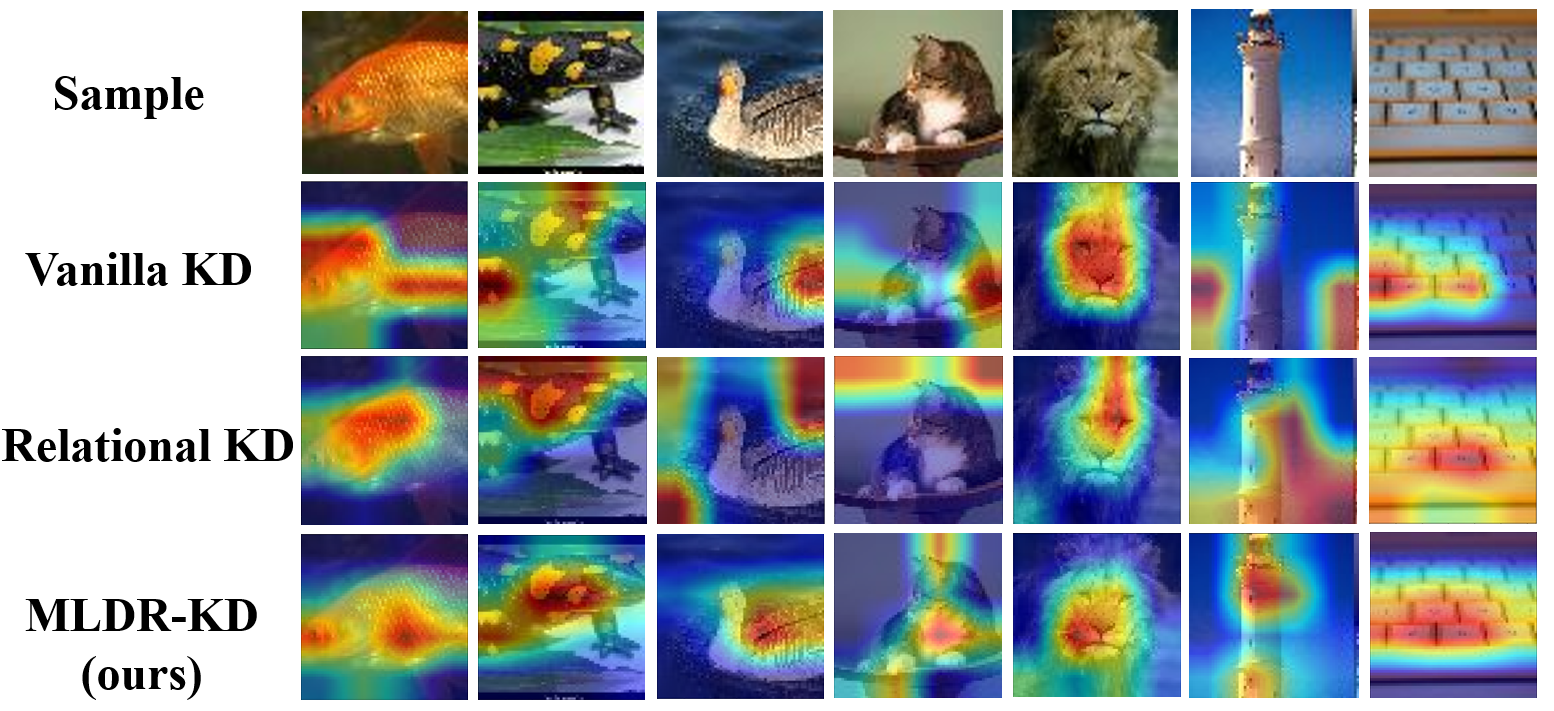}
  \vspace{-3mm}
  \caption{Comparisons of feature visualizations when using kinds of knowledge distillation methods. The teacher is Vision Mamba Tiny~\cite{vim}, the student is ResNet-18~\cite{he2016deep}. The direct use of conventional relational KD underperforms on heterogeneous distillation, while our MLDR-KD could greatly improve this problem.}
  \label{fig:featuremap}
  \vspace{-5mm}
\end{figure}

A few recent studies attempt to investigate the feasibility of using heterogeneous teachers for knowledge transfer~\cite{yu2024unleashing,Liu2024DDKDD,wang2024the}. Touvron \textit{et al.}~\cite{touvron2022deit} achieves successful training of a ViT student model using a CNN teacher model. Ao Wang \textit{et al.}~\cite{wang2024repvit} revisits the efficient design of lightweight CNNs from the ViT perspective and emphasizes their promising prospect for mobile devices. Although achieving good results, these approaches cannot be extended to various architectures.
As a pioneer, Zhiwei Hao \textit{et al.}~\cite{hao2024one} finds there is a huge gap among feature maps of heterogeneous architecture, resulting in the failure of feature-based knowledge distillation~\cite{peng2019correlation,Heo_2019_ICCV,zagoruyko2022paying,romero2014fitnets,Heo2018KnowledgeTV,Yim2017AGF,Ahn2019VariationalID,Chen2022KnowledgeDW,Chen2021DistillingKV,guo2023class,Li2021OnlineKD,Lin2022KnowledgeDV}. Thus, they propose logit-based generic heterogeneous distillation. 
Specifically, by increasing the confidence in the correct category of the teacher model, the impact of architectural differences is reduced, and the results are improved. However, this approach somehow weakens the transfer of dark knowledge, which is regarded as very important in knowledge distillation (e.g., whether a sample that is actually a dog or more like a cat), which limits the performance of heterogeneous distillation.

In this paper, we further explore how to effectively transfer the dark knowledge during heterogeneous distillation for the first time. In traditional homogeneous distillation, relational knowledge distillation (RKD)~\cite{park2019relational} is generally considered as an effective method for transferring dark knowledge, as shown in Fig.~\ref{fig:header}. RKD aligns correlations or dependencies among multiple instances between the student and teacher networks. 
However, we find that the direct use of RKD in heterogeneous distillation causes a new problem: the over-amplification of the role of dark knowledge, which may reduce the confidence in the correct category of the teacher model. Since the latter is equally important in heterogeneous distillation due to the variability between architectures, this can directly contribute to the failure of the RKD method, as shown in Fig.~\ref{fig:featuremap}. Facing such a dilemma, a question naturally arises: \textit{can we effectively transfer the abundant dark knowledge while keeping the confidence of the correct category during heterogeneous distillation?} 

To answer this question, we present an innovative framework called Multi-Level Decoupled Relational Knowledge Distillation (MLDR-KD) for heterogeneous distillation. Specifically, we first propose Decoupled Finegrained Relation Alignment (DFRA),
in which model logits are first decoupled into multiple finegrained relationships between different categories under each image and different images under each category. Due to the multiple steps finegrained decoupling, the subsequent alignment is sensitive to whether the model classifies correctly, and it can magnify the gap when the classification results of student model and teacher model are not aligned.
As a result, our method can well transfer dark knowledge while enhancing the confidence of the classification results during heterogeneous distillation. Further, we apply the DFRA to both logit and feature levels, and 
present the Multi-Scale Dynamic Fusion (MSDF) module in the feature level.
In the MSDF module, the multiscale feature maps of different stages in student model are projected into multiple logits, 
and a gated network is used to dynamically fuse these logits.
As shown in Fig.~\ref{fig:featuremap}, our method can release the potential of logit-based cross-architectures distillation, where the student model will focus more on information related to the goal.

To illustrate the robustness and generality of our approach, we conduct 12 kinds of experiments between 4 architectures including CNNs, Transformers, MLPs and Mambas. We distill them two by two, with image classification as the evaluation task and acc@1 as the evaluation metric. Compared with the best available method, our MLDR-KD framework improves student model performance with gains of up to 1.43\%, 4.86\%, 0.93\%, 0.83\% on CIFAR-100 dataset and 1.57\%, 2.78\%, 1.61\%, 2.13\% on Tiny-ImageNet dataset for CNNs, Transformers, MLPs and Mambas architectures under the same conditions, respectively. The ablation study has also demonstrated the effectiveness of our methods.
In summary, our main contributions can be summarized as follows:
\begin{enumerate}
    \item[$\bullet$] We first propose to utilize dark knowledge for heterogeneous distillation. We find that: 1) previous work~\cite{hao2024one} destroys the dark knowledge present in the teacher model logit, which limits the performance of heterogeneous distillation; 2) The direct use of relational knowledge distillation in traditional homogenous distillation 
    to transfer dark knowledge 
    reduces the confidence in the correct category, bringing about catastrophic performance in heterogeneous distillation. 

    \item[$\bullet$] To address these, we present a novel framework called Multi-Level Decoupled Relational Knowledge Distillation (MLDR-KD).
    It consists of Decoupled Finegrained Relation Alignment (DFRA) and Multi-Scale Dynamic Fusion (MSDF) module. Specifically, DFRA enables the student model to learn more finegrained relationships in both logit and feature levels. MSDF module further improves the feature level DFRA by dynamically fusing the predictions of multiscale features of students.

    \item[$\bullet$] Extensive experiments across diverse datasets and models consistently verify that MLDR-KD can achieve new state-of-the-art performance. In particular, we extend the MLDR-KD method to the new architecture Mamba, and find our method also performs best, which well illustrates the robustness and generality of our method.
\end{enumerate}

%% file: sec/2_related_work.tex
\section{Related work}
\label{sec:related_work}

\textbf{Homogeneous Distillation}
Hinton \textit{et al.}~\cite{hinton2015distilling} firstly introduces knowledge distillation to transfer a teacher’s knowledge to a student by minimizing their Kullback-Leibler divergence. Following works can be mainly categorized into two pipelines: Feature-based KD and Logits-based KD. To enhance representational capacity, Feature-based KD methods~\cite{romero2014fitnets,park2019relational} distill knowledge from both intermediate layers and logit outputs. Subsequent works explore various perspectives: CRD~\cite{tian2019contrastive} emphasizes the structural knowledge of the teacher, while CC~\cite{peng2019correlation} identifies instance-level congruent constraints, transferring both instance-level information and inter-instance correlation. Further advancements~\cite{guo2023class,yang2022masked,yang2022focal} refine this process with class activation mapping, feature masking, and focal techniques for object detection. Logits-based KD enhances student models by transferring softened targets from teacher models~\cite{hinton2015distilling,zhao2022decoupled}.~\cite{sun2024logit} introduces a Z-score logit standardization method to better capture inter-logit relations to conquer the shared-temperature constraint. 

However, in logits-based KD simply using KL divergence is insufficient for exact matching. To tackle the issue,~\cite{huang2022knowledge} proposes a relation-based loss to preserve inter-class relationships.~\cite{yang2022crossimagerelationalknowledgedistillation} proposes a novel Cross-Image Relational KD (CIRKD), which focuses on transferring structured pixel-to-pixel and pixel-to-region relations among the whole images.~\cite{guo2023linklesslinkpredictionrelational} proposes a relational KD framework, Linkless Link Prediction (LLP), to distill knowledge for link prediction with MLPs. These methods seem to solve the problem that dark knowledge is not well transferred in heterogeneous distillation.
Nonetheless, these relational distillation methods 
smooth the logit too much, leading to a reduction in the confidence in the correct category. Moreover, we show that directly transferring conventional relational KD to the heterogeneous distillation setting proves ineffective. Thus, it is a significant necessity to investigate the effective application of relational distillation methods in heterogeneous distillation. 

\textbf{Heterogeneous Distillation}
Heterogeneous distillation allows efficient models to inherit rich representations from powerful teacher models of different architectures, enhancing student model performance and generalization across architectural boundaries. Liu \textit{et al.}~\cite{liu2022cross} pioneers heterogeneous knowledge distillation by aligning the output, attention, and feature spaces of heterogeneous models, assuming identical pixel-level spatial information. To overcome the limitations of this assumption,~\cite{zhao2023cross} addresses the architecture gap in cross-architecture distillation by synchronizing the pixel-wise receptive fields of teacher and student networks. However, these methods overlook spatial differences and global context, which FASD~\cite{yu2024unleashing} addresses by aligning heterogeneous features and logit mappings between Transformer and Mamba models. However, these approaches do not directly scale to all heterogeneous architectures.
Furthermore, OFA-KD~\cite{hao2024one} explores the feasibility of distilling between multiple architectures. They identified two key limitations in existing methods: lack of latent space alignment, causing inconsistencies in heterogeneous distillation, and absence of adaptive target enhancement, weakening focused knowledge transfer. OFA-KD introduces latent space alignment to eliminate architecture-specific information and adaptive target enhancement to sharpen knowledge transfer, achieving notable gains across diverse models.
In this paper, we find that dark knowledge is severely corrupted as OFA-KD changes the distribution of the output logit of the teacher model, which limits the performance of heterogeneous distillation. 
Therefore, we design a novel heterogeneous relational KD framework called MLDR-KD, which can retain redundant dark knowledge while enhancing confidence in the correct target.

\begin{figure*}[t]
  \centering
  \includegraphics[width=0.96\linewidth]{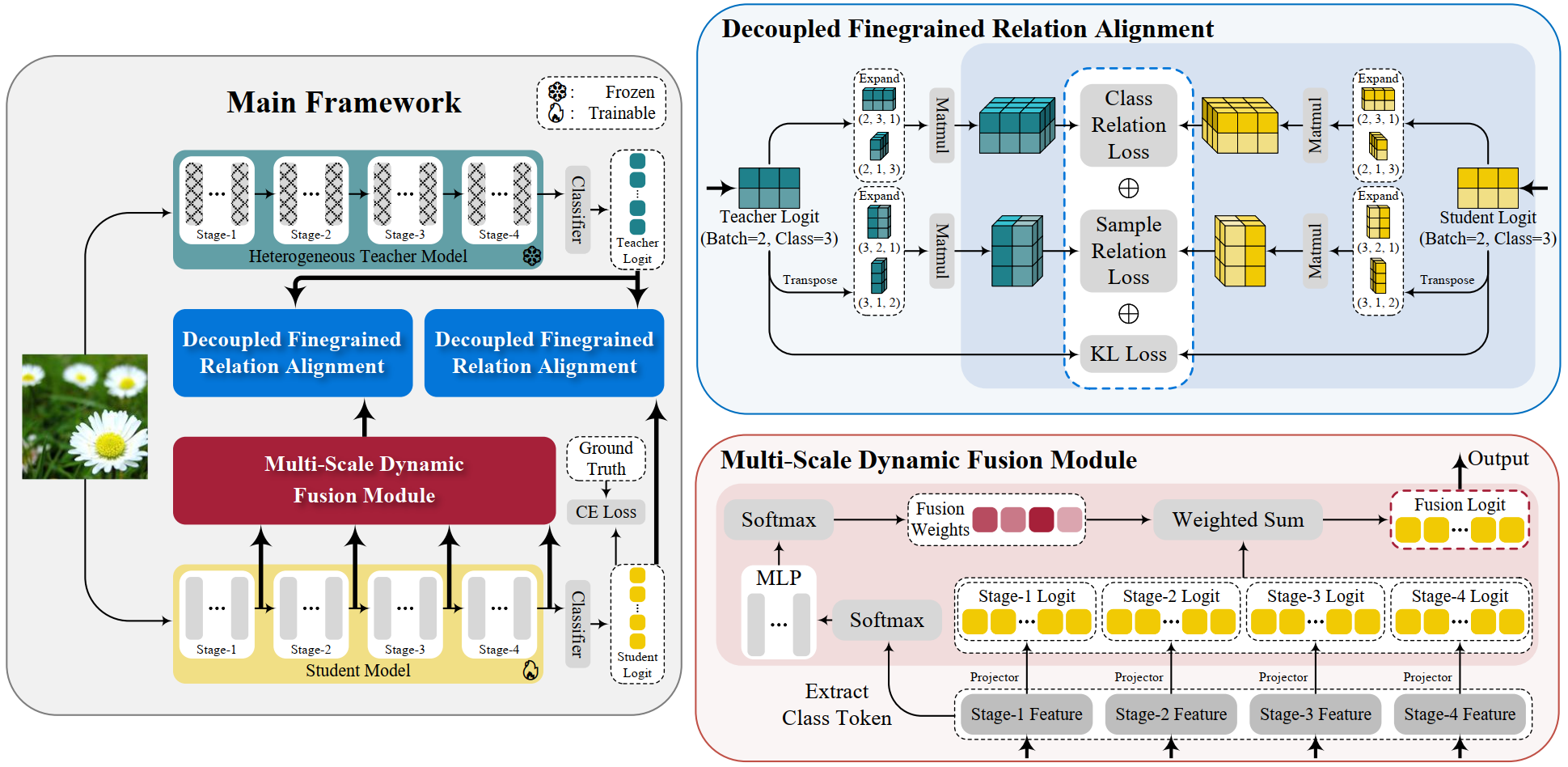}
  \vspace{-3mm}
  \caption{Overview of the proposed MLDR-KD framework. It comprises two main components:  Decoupled Finegrained Relation Alignment (DFRA), and Multi-Scale Dynamic Fusion (MSDF). In DFRA, after obtaining the logits of teacher and student, we decouple them into class-wise relation and sample-wise relation, and then align these relationships via Kullback-Leibler divergence. DFRA is applied to both logit and feature levels. MSDF further improves the effect of feature-level DFRA by dynamically fusing feature maps of student.
  }
  \vspace{-5mm}
  \label{fig:framework}
\end{figure*}

%% file: sec/3_methodology.tex
\section{Methodology}
\label{sec:methodology}
\subsection{Preliminaries}
\label{sec:KD}
We start from the original Logit-based Knowledge Distillation (KD) method. Generally, We denote the logit out as $z \in \mathbb{R}^{B\times N}$, where $B$ is the batch size in training and $N$ means the number of categories in dataset. The softmax function is then used to obtain a probability distribution:

\begin{equation}
    \begin{aligned}
        p_{i} = \frac{\exp(z_{i})}{\sum_{j = 1}^{N}\exp(z_{j})}, i = 1,2,\cdots,N
    \end{aligned}
\end{equation}
where $p_{i}$ is the probability distribution of a sample.

In Logit-based KD, the cross-entropy loss $\mathcal{L}_{CE}$ is used to minimize gap between the student model and the ground truth: 
\begin{equation}
    \begin{aligned}
        \mathcal{L}_{CE}=-\sum_{i = 1}^{B}\sum_{j = 1}^{N}y_{ij}\log(p_{s,ij})
    \end{aligned}
\end{equation}
where $y_{ij}$ is the one-hot encoded true label, and $p_{s,ij}$ is the probability distribution of the student model after softmax.

Student model mimics the teacher model by means of distillation loss $\mathcal{L}_{KL}$. We use the Kullback-Leibler divergence to measure the difference between the student model and the teacher model: 
\begin{equation}
    \begin{aligned}
        \mathcal{D}_{KL}(p_{s,i}||p_{t,i})=\sum_{j = 1}^{N}p_{s,ij}\log\frac{p_{s,ij}}{p_{t,ij}}
    \end{aligned}
\end{equation}
\begin{equation}
    \begin{aligned}
        \mathcal{L}_{KL}=\frac{1}{B}\sum_{i = 1}^{B}\mathcal{D}_{KL}(p_{s,i}||p_{t,i})
    \end{aligned}
\end{equation}
where $\mathcal{D}_{KL}(p_{s,i}||p_{t,i})$ is the KL divergence. 

The overall knowledge distillation loss function: 
\begin{equation}
    \begin{aligned}
        \mathcal{L} = \mathcal{L}_{CE}+\lambda \mathcal{L}_{KL}
    \end{aligned}
\end{equation}
where $\lambda$ is a weighting parameter that balances the cross-entropy loss and distillation loss.

\subsection{MLDR-KD}
The overview of MLDR-KD is depicted in Fig.~\ref{fig:framework}. Our method framework is primarily divided into two modules: the Decoupled Finegrained Relation Alignment (DFRA) and the Multi-Scale Dynamic Fusion (MSDF) Module. Initially, the student model are segmented into multiple stages. After forward inference, the logit outputs of the teacher and the student are obtained. Meanwhile, feature maps of student at each stage are fused by MSDF Module to get a fusion logit. Finally, the fusion logit and the logit output of student will be aligned with logit output of teacher by DFRA in logit and feature levels. We present our two modules of MLDR-KD framework in Sec.~\ref{sec:DFRA} and Sec.~\ref{sec:MSDFM}.

\subsubsection{Decoupled Finegrained Relation Alignment}

\begin{figure*}[t]
  \centering
  \includegraphics[width=0.98\linewidth]{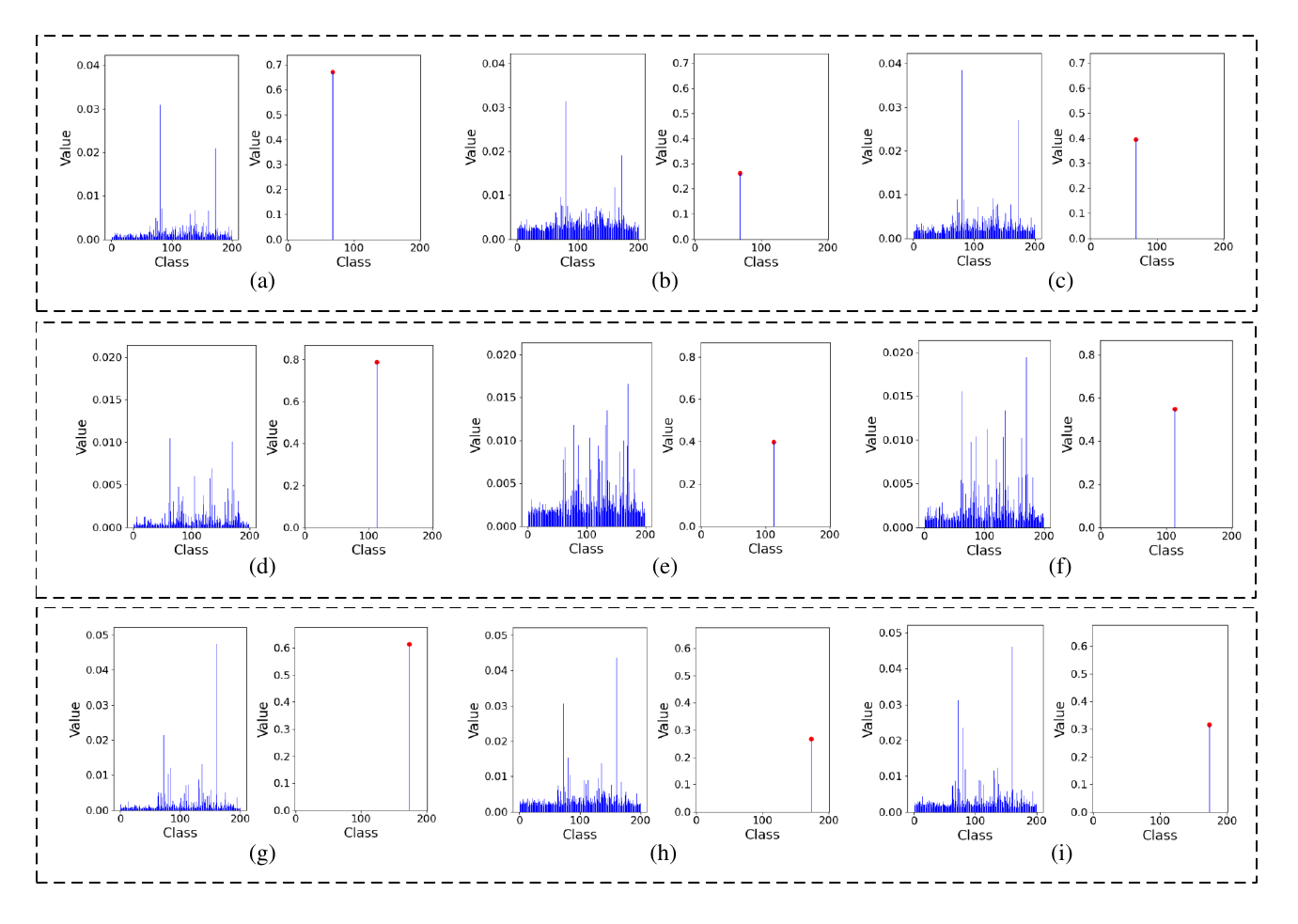}
  \vspace{-6mm}
  \caption{Comparisons of the averaged prediction distribution of all samples of single category among OFA-KD ((a),(d),(g)), RKD ((b),(e),(h)), and our MLDR-KD ((c),(f),(i)). Three black boxes represent three randomly selected categories.
  In each figure (left), we show the logit of category in addition to the correct category. In each figure (right), the logit of the correct category is displayed. From  the figure we can see that our method has 
high confidence for the correct category while transferring abundant dark knowledge in the teacher model logit.
  }
  \label{fig:logit}  
  \vspace{-4mm}
\end{figure*}

\label{sec:DFRA}
In heterogeneous distillation, it is crucial to balance the correct samples' confidence and the dark knowledge from teacher model. To deal with this problem, we propose DFRA to enhance knowledge transfer between heterogeneous architectures. As shown in Fig.~\ref{fig:framework}, we decouple logit prediction into Class-Wise Relation and Sample-Wise Relation in contrast to exact match. These relations will be aligned in multi level. 

\textbf{Class-Wise Relation Decoupling} \ Class-Wise relation represents the degree of similarity among different categories. In this section, we refine this relationship to each sample in the batch to transfer more information (\emph{e.g.} under a particular sample labeled dog, the similarity between cat and elephant). Firstly, we expand logit prediction to three dimensions, which is defined as:
\begin{equation}
    \begin{aligned}
        \hat{z}_{c} = \operatorname{Expand}(z/T), \hat{z}_{c} \in \mathbb{R}^{B\times N\times 1}
    \end{aligned}
\end{equation}
where $T$ is the soft factor in knowledge distillation. 
Then we could calculate its self-relation, which is implemented as the scaled product relation:
\begin{equation}
    \begin{aligned}
        \mathcal{R}_{class} =\operatorname{Softmax} \left(\frac{\hat{z}_{c} \hat{z}_{c}^{T}}{\sqrt{N}}\right), \mathcal{R}_{class} \in \mathbb{R}^{B\times N\times N}
    \end{aligned}
\end{equation}
where $\mathcal{R}_{class}$ indicates class-wise relation decoupled from initial logit out $z$. $N$ denotes a scaling factor that equals to the number of categories in dataset.

\textbf{Sample-Wise Relation Decoupling} \ The other information then can be decoupled from initial logit out $z$ is sample-wise relation. It's regarded as the degree of similarity between samples under one category (\emph{e.g.} in a batch which of the many samples is more like a dog). Sample-wise relation can be modeled by predictions of a batch of data as follows:
\begin{equation}
    \begin{aligned}
        \hat{z}_{b} = \operatorname{Expand}(z^{T}/T), \hat{z}_{b} \in \mathbb{R}^{N\times B\times 1}
    \end{aligned}
\end{equation}
\begin{equation}
    \begin{aligned}
        \mathcal{R}_{sample} = \operatorname{Softmax}\left(\frac{\hat{z}_{b} \hat{z}_{b}^{T}}{\sqrt{N}}\right), \mathcal{R}_{batch} \in \mathbb{R}^{N\times B\times B}
    \end{aligned}
\end{equation}
where $\mathcal{R}_{sample}$ indicates sample-wise relation decoupled from initial logit out $z$.

\textbf{Multiple Relation Alignment} \ Decoupled finegrained relation between heterogeneous student model and teacher model can be aligned by Kullback-Leibler divergence:
\begin{equation}
    \begin{aligned}
        \mathcal{L}_{class} = \mathcal{L}_{KL}(\mathcal{R}_{class}^{s},\mathcal{R}_{class}^{t})
    \end{aligned}
\end{equation}
\begin{equation}
    \begin{aligned}
        \mathcal{L}_{sample} = \mathcal{L}_{KL}(\mathcal{R}_{sample}^{s},\mathcal{R}_{sample}^{t})
    \end{aligned}
\end{equation}
\begin{equation}
    \begin{aligned}
        \mathcal{L}_{total} = \mathcal{L}_{class}+\mathcal{L}_{sample}+\lambda\mathcal{L}_{KL}(p^{s},p^{t})
    \end{aligned}
\end{equation}
where $\lambda$ denotes the balance coefficient. $p^{s}$ and $p^{t}$ are the probability distributions of $z^{s}$ and $z^{t}$ after softmax function.

\begin{table*}[t]
\centering
\caption{Results on CIFAR100 dataset. The best results are indicated in bold. For the baseline, most of the experimental results are inherited from OFAKD, while the additional experiments we conducted are marked with *.}
\vspace{-2mm}
\label{tab:cifar100}
\resizebox{1.7\columnwidth}{!}{%
\fontsize{7pt}{7}\selectfont
\begin{tabular}{@{}cccccccccc@{}}
\toprule
Student Model        & From Scratch & Teacher Model & From Scratch & KD~\cite{hinton2015distilling}    & RKD~\cite{park2019relational}   & DKD~\cite{zhao2022decoupled}   & OFAKD~\cite{hao2024one}  & \textbf{MLDRKD} & {\color[HTML]{FF0000} $\Delta$}     \\ \midrule
\textit{CNNs-based}   &              &               &              &       &       &       &        &                 & {\color[HTML]{FF0000} }      \\ \midrule
                     &              & ViT-S         & 92.04        & 77.26 & 73.72 & 78.10 & 80.15  & \textbf{80.51}  & {\color[HTML]{FF0000} +0.36} \\
ResNet18             & 74.01        & Swin-T        & 89.26        & 78.74 & 74.11 & 80.26 & 80.54  & \textbf{81.56}  & {\color[HTML]{FF0000} +1.02} \\
                     &              & Mixer-B/16    & 87.29        & 77.79 & 73.75 & 78.67 & 79.39  & \textbf{80.79}  & {\color[HTML]{FF0000} +1.40}  \\
                     &              & ViM-S         &  87.89*       & 78.22*     & 77.41*     & 79.20*     & 79.90*  & \textbf{80.23}  & {\color[HTML]{FF0000} +0.33} \\ \midrule
MobileNetV2          & 73.68        & ViT-S         & 92.04        & 72.77 & 68.46 & 69.80 & 78.45  & \textbf{79.31}  & {\color[HTML]{FF0000} +0.86} \\
                     &              & Mixer-B/16    & 89.26        & 73.33 & 68.95 & 70.20 & 78.78  & \textbf{80.21}  & {\color[HTML]{FF0000} +1.43} \\ \midrule
\textit{Transformers-based}   &              &               &              &       &       &       &        &                 & {\color[HTML]{FF0000} }      \\ \midrule
                     &              & Mixer-B/16    & 87.29        & 75.93 & 69.89 & 76.39 & 78.93  & \textbf{80.09}  & {\color[HTML]{FF0000} +1.16} \\
Swin-P               & 72.63        & ConvNeXt-T    & 88.41        & 76.44 & 69.79 & 76.80 & 78.32  & \textbf{81.21}  & {\color[HTML]{FF0000} +2.89} \\
                     &              & ViM-S         & 87.89*       & 78.42*     &  72.69*      & 79.29*    & 79.48* & \textbf{79.91}  & {\color[HTML]{FF0000} +0.43} \\ \midrule
                     &              & Mixer-B/16    & 87.29        & 71.36 & 70.82 & 73.44 & 73.90  & \textbf{78.76}  & {\color[HTML]{FF0000} +4.86} \\
Deit-T               & 68.00        & ConvNeXt-T    & 88.41        & 72.99 & 71.73 & 74.60 & 75.76  & \textbf{79.18}  & {\color[HTML]{FF0000} +3.42} \\
                     &              & ViM-S         & 87.89*       & 73.28*     & 70.22*     & 74.68*     & 76.69* & \textbf{77.27}  & {\color[HTML]{FF0000} +0.58} \\ \midrule
                     &              & Mixer-B/16    & 87.29*       & 77.43*     & 75.76*     & 79.53*     & 81.54* & \textbf{81.61}  & {\color[HTML]{FF0000} +0.07} \\
T2t ViT-7            & 74.74        & ConvNeXt-T    & 88.41*       & 79.26*     & 75.31*     &  79.83*     & 82.52* & \textbf{82.67}  & {\color[HTML]{FF0000} +0.15} \\
                     &              & ViM-S         & 87.89*       & 77.39*     & 72.53*     & 78.48*     & 81.38* & \textbf{81.47}  & {\color[HTML]{FF0000} +0.09} \\ \midrule
\textit{MLPs-based}   &              &               &              &       &       &       &        &                 & {\color[HTML]{FF0000} }      \\ \midrule
                     &              & ConvNeXt-T    & 88.41        & 72.25 & 65.82 & 73.22 & 81.22  & \textbf{81.96}  & {\color[HTML]{FF0000} +0.74} \\
ResMLP-S12           & 66.56        & Swin-T        & 89.26        & 71.89 & 64.66 & 72.82 & 80.63  & \textbf{81.56}  & {\color[HTML]{FF0000} +0.93} \\
                     &              & ViM-S         & 87.89*       & 80.23*     & 78.19*     & 80.72*     & 80.37* & \textbf{80.92}  & {\color[HTML]{FF0000} +0.20} \\ \midrule
\textit{Mambas-based} &              &               &              &       &       &       &        &                 & {\color[HTML]{FF0000} }      \\ \midrule
                     &              & ViT-S         & 92.04*       & 77.55*     & 68.85*     & 79.58*     & 81.24* & \textbf{81.87}  & {\color[HTML]{FF0000} +0.63} \\
ViM-T                & 70.99        & Swin-T        & 89.26*       & 78.53*     & 66.91*     & 80.50*      & 82.22* & \textbf{83.01}  & {\color[HTML]{FF0000} +0.79} \\
                     &              & Mixer-B/16    & 87.29*       & 79.34*     & 73.08*     & 80.57*     & 82.19* & \textbf{83.02}  & {\color[HTML]{FF0000} +0.83} \\
                     &              & ConvNeXt-T    & 88.41*       &  80.59*     &  66.41*     & 82.51*     & 82.89* & \textbf{82.95}  & {\color[HTML]{FF0000} +0.06} \\ \bottomrule
\end{tabular}%
}
\vspace{-4mm}
\end{table*}

Due to the multi-step decoupling of logit, our proposed Decoupled Finegrained Relation Alignment method is robust to enhance the performance of student model. DFRA not only improves the confidence level of the classification results, but also retains a lot of details of the information. Following experiments will further demonstrate the effectiveness of our method.
\subsubsection{Multi-Scale Dynamic Fusion Module}
\label{sec:MSDFM}
In heterogeneous distillation, it can transfer more knowledge in addition to the logit level. Specifically, there is a huge gap in the feature maps between heterogeneous models. So it seems feasible to transfer feature-level knowledge in a latent logit space. In other words, feature maps at each stage of student are projected to logit space to be aligned. It can be viewed as training each stage as a separate student. However, the learning abilities of the student models at different stages are disparate because they have different numbers of parameters. It is manifestly inappropriate to assign them the same weighting for learning. So we propose a method that introduce class token in each stage of the student model, denoted as $x_{i}$ where $i$ is ordinal number of each stage, to dynamically balance weighting of each stage.

The student model is divided into four stages, each of which requires feature matching with the teacher model in logit space, as shown in Figure 2. For each forward inference, the student model outputs features of four stages $\{f_{i}\}_{i=1}^{4}$. We split $\{f_{i}\}_{i=1}^{4}$ into the class token $\{x_{i}\}_{i=1}^{4}$ for each stage and the architecture-independent feature information $\{\hat{f}_{i}\}_{i=1}^{4}$. After that, $\{\hat{f}_{i}\}_{i=1}^{4}$ is mapped to the logit space through the projector, denoted as $\{\hat{p}_{i}\}_{i=1}^{4}$. We use the global semantic information contained in the class token $\{x_{i}\}_{i=1}^{4}$ at each stage to dynamically balance the feature matching under logit space. We apply an MLP layer to generate the balancing weights. MLP layer can be represented as follows:
\begin{equation}
    \begin{aligned}
         X_{token} &= \operatorname{Stack}(\{x_{i}\}_{i=1}^{4}) \\
         X_{hidden} &= \operatorname{GELU}(\operatorname{Linear}(X_{token}))   \\
         W_{balance} &= \operatorname{Softmax}(\operatorname{Linear}(X_{hidden}))
    \end{aligned}
\end{equation}
where $\operatorname{Stack}(\cdot)$ denotes a stacking function for class token aggregation. $\operatorname{GELU}(\cdot)$ indicates an activation function. $\operatorname{Linear}(\cdot)$ is a fully connected layer. 
In MLP layer, the token sequence $\{x_{i}\}_{i=1}^{4}$ is compressed into the vector $X_{token}$. Then with a linear layer and softmax function, we can calculate the balancing weights $W_{balance}$. Further, we use dot product to balance the $\{\hat{p}_{i}\}_{i=1}^{4}$, as
\begin{equation}
    \begin{aligned}
        P_{stage} &= \operatorname{Stack}(\{\hat{p}_{i}\}_{i=1}^{4})\\
        Logit &= W_{balance}\cdot P_{stage}
    \end{aligned}
\end{equation}
where $Logit$ is logit output balanced by class token. Finally, we employ DFRA in Sec.~\ref{sec:DFRA} to minimize the gap between $Logit$ and teacher logit $p^{t}$.
\begin{equation}
    \begin{aligned}
        \mathcal{L}_{balance} = \operatorname{DFRA}(Logit, p^{t})
    \end{aligned}
\end{equation}

\subsubsection{Effectiveness Analysis}
In Fig.~\ref{fig:logit}, we compare the averaged prediction distribution of all samples of a single category among OFA-KD ~\cite{hao2024one}, RKD
~\cite{park2019relational}, and our MLDR-KD. Three categories are randomly selected.
By comparing each figure left in Fig.~\ref{fig:logit}, we can find that conventional RKD and our MLDR-KD both retain more dark knowledge than the previous heterogeneous distillation method OFA-KD.
However, as each figure right shows, conventional RKD reduces the confidence of the student model in the correct category, which leads to its poor performance in heterogeneous distillation. 
In contrast, the student model trained by our MLDR-KD has high confidence for the correct category while transferring abundant dark knowledge in the teacher model logit, which is consistent with our key observations.

%% file: sec/4_experiments.tex

\begin{table*}[t]
\centering
\caption{Results on Tiny-ImageNet dataset. The best results are indicated in bold. We conducted all of the additional experiments in baseline. CNN-based experiments are through 100 epochs training. Other experiments are through 300 epochs training. }
\vspace{-2mm}
\label{tab:tinyimg}
\resizebox{1.4\columnwidth}{!}{%
\fontsize{7pt}{7}\selectfont
\begin{tabular}{@{}cccccccc@{}}
\toprule
Student Model        & From Scratch & Teacher Model                & From Scratch                 & KD                      & OFAKD                        & \textbf{MLDRKD}                       & {\color[HTML]{FF0000} $\Delta$}     \\ \midrule
\textit{CNNs-based}   &              &                              &                              &                         &                              & \textbf{}                             & {\color[HTML]{FF0000} }      \\ \midrule
                     &              & ViT-S                        & 80.03                        & 65.34                   & 65.82                        & \textbf{67.13}                        & {\color[HTML]{FF0000} +1.31} \\
ResNet18             & 63.39        & Swin-T                       & 76.13                        & 66.20                   & 66.94                        & \textbf{68.51}                        & {\color[HTML]{FF0000} +1.57} \\
                     &              & Mixer-B/16                   & 69.74                        & 64.42                   & 65.03                        & \textbf{66.02}                        & {\color[HTML]{FF0000} +0.99} \\
                     &              & ViM-T                        & 76.13                        & 66.69                   & 66.62                        & {\color[HTML]{000000} \textbf{67.61}} & {\color[HTML]{FF0000} +0.92} \\ \midrule
                     &              & ViT-S                        & 80.03                        & 66.00                   & 65.56                        & \textbf{66.96}                        & {\color[HTML]{FF0000} +0.96} \\
MobileNetV2          & 63.93        & Swin-T                       & 76.13                        & 66.51                   & 66.60                        & \textbf{68.06}                        & {\color[HTML]{FF0000} +1.46} \\
                     &              & Mixer-B/16                   & 69.74                        & 64.89                   & 65.28                        & \textbf{65.54}                        & {\color[HTML]{FF0000} +0.26} \\
                     &              & ViM-T                        & 76.13                        & 66.26                   & 66.14                        & {\color[HTML]{000000} \textbf{67.24}} & {\color[HTML]{FF0000} +0.98} \\ \midrule
\textit{Transformers-based}   &              &                              &                              &                         &                              & \textbf{}                             & {\color[HTML]{FF0000} }      \\ \midrule
                     &              & Mixer-B/16                   & 69.74                        & 68.67                   & 68.24                        & \textbf{69.10}                        & {\color[HTML]{FF0000} +0.43} \\
Swin-P               & 65.09        & ConvNeXt-T                   & 72.82                        & 66.90                   & 67.74                        & \textbf{68.03}                        & {\color[HTML]{FF0000} +0.29} \\
                     &              & ResNet50                     & 74.61                        & 70.84                   & 71.90                        & \textbf{72.36}                        & {\color[HTML]{FF0000} +0.46} \\
                     &              & {\color[HTML]{000000} ViM-T} & 76.13                        & 70.63                   & 70.22                        & {\color[HTML]{000000} \textbf{70.83}} & {\color[HTML]{FF0000} +0.20} \\ \midrule
                     &              & Mixer-B/16                   & 69.74                        & 64.13                   & 68.74                        & \textbf{69.26}                        & {\color[HTML]{FF0000} +0.52} \\
                     &              & ConvNeXt-T                   & 72.82                        & 59.33                   & 62.83                        & \textbf{64.86}                        & {\color[HTML]{FF0000} +2.03} \\
Deit-T               & 58.27        & ResNet50                     & 74.61                        & 66.72                   & 71.89                        & \textbf{72.29}                        & {\color[HTML]{FF0000} +0.4}  \\
                     &              & ViM-S                        & 83.86                        & 66.19                   & 66.96                        & \textbf{68.44}                        & {\color[HTML]{FF0000} +1.48} \\
                     &              & ViM-T                        & 76.13                        & 67.56                   & 68.69                        & \textbf{71.47}                        & {\color[HTML]{FF0000} +2.78} \\ \midrule
                     &              & Mixer-B/16                   & 69.74                        & 67.34                   & 68.85                        & \textbf{69.36}                        & {\color[HTML]{FF0000} +0.51} \\
T2t ViT-7            & 64.37        & ConvNeXt-T                   & 72.82                        & 65.16                   & 66.65                        & \textbf{69.31}                        & {\color[HTML]{FF0000} +2.66} \\
                     &              & ResNet50                     & 74.61                        & 70.08                   & 70.41                        & \textbf{72.66}                        & {\color[HTML]{FF0000} +2.25} \\
                     &              & ViM-T                        & 76.13                        & 69.89                   & 70.79                        & {\color[HTML]{000000} \textbf{72.22}} & {\color[HTML]{FF0000} +1.43} \\ \midrule
\textit{MLPs-based}   &              &                              &                              &                         &                              & \textbf{}                             & {\color[HTML]{FF0000} }      \\ \midrule
                     &              & ConvNeXt-T                   & 72.82                        & 66.37                   & 66.74                        & \textbf{67.23}                        & {\color[HTML]{FF0000} +0.49} \\
                     &              & ResNet50                     & 74.61                        & 72.06                   & 70.63                        & \textbf{73.44}                        & {\color[HTML]{FF0000} +1.38} \\
ResMLP-S12           & 65.46        & {\color[HTML]{000000} ViM-T} & {\color[HTML]{000000} 76.13} & 71.58                   & {\color[HTML]{000000} 70.31} & {\color[HTML]{000000} \textbf{71.72}} & {\color[HTML]{FF0000} +0.14} \\
                     &              & Swin-T                       & 76.13                        & 71.70                   & 73.09                        & \textbf{73.21}                        & {\color[HTML]{FF0000} +0.12} \\
                     &              & ViT-S                        & 80.03                        & 70.32                   & 69.64                        & \textbf{71.93}                        & {\color[HTML]{FF0000} +1.61} \\ \midrule
\textit{Mambas-based} &              &                              &                              &                         &                              & \textbf{}                             & {\color[HTML]{FF0000} }      \\ \midrule
                     &              & ViT-S                        & 80.03                        & 67.66                   & 72.84                        & \textbf{74.97}                        & {\color[HTML]{FF0000} +2.13} \\
ViM-T                & 61.85        & Swin-T                       & 76.13                        & 70.53                   & 72.08                        & \textbf{73.31}                        & {\color[HTML]{FF0000} +1.23} \\
                     &              & Mixer-B/16                   & 69.74                        & 65.59                   & 69.63                        & \textbf{70.55}                        & {\color[HTML]{FF0000} +0.92} \\ \bottomrule
\end{tabular}%
}
\vspace{-4mm}
\end{table*}

\section{Experiments}
\label{sec:experiments}
\subsection{Dataset and Settings}

\par In this section, we will introduce the dataset used in the experiment
and the implementation details.

\textbf{Datasets} \ We validate the proposed method on CIFAR-100~\cite{krizhevsky2009learning} and Tiny-ImageNet~\cite{le2015tiny}. The CIFAR-100 dataset comprises 60,000 images divided into 100 categories, with 600 images per category. The size of each image is 32×32. 50,000 images are used as the training set, and 10,000 are used as the test set. The Tiny-ImageNet dataset is a smaller version of the ImageNet dataset. It contains 100,000 images, which are divided into 200 categories. Each category has 500 training images, 50 validation images, and 50 test images. Each image is resized to 64×64.

\textbf{Implementation Details} \
To validate the generality of our method, we conduct experiments with different student and teacher models. For student models, CNN-based ResNet18~\cite{he2016deep}, MobileNet-v2~\cite{howard2018inverted}, Transformers-based Swin-p~\cite{liu2021swin}, Deit-t~\cite{touvron2021training}, T2t Vit-7~\cite{yuan2021tokens}, MLP-based ResMLP-S12~\cite{touvron2022resmlp}, and Mamba-based Vim-t~\cite{vim}, are selected. For teacher models, Resnet50~\cite{he2016deep}, Vit-S~\cite{dosovitskiy2020image}, Swin-T~\cite{liu2021swin}, Mixer-B/16~\cite{tolstikhin2021mlp}, and ConvNeXt-T~\cite{liu2022convnet}, are considered.
For CNNs, the SGD is adopted as the optimizer, with a base learning rate of 0.05. For Transformers, MLPs, and Mambas, the Adamw~\cite{loshchilovdecoupled} is adopted as the optimizer, with a base learning rate of 5e-4. The cosine learning rate decay strategy is used. For all datasets, we set the batch size as 128.
The training epoch number of CIFAR-100 is 300 for all models. For Tiny-ImageNet, CNNs are trained with 100 epochs, whereas ViTs, MLPs, and Mambas are trained with 300 epochs. All experiments are conducted using Nvidia RTX 3090 GPU.

\begin{table*}[htbp]
\centering
\begin{minipage}[t]{0.22\linewidth}
\centering
\captionof{table}{Impact of the number of stages in MSDF (Multi-Scale Dynamic Fusion ).}
\vspace{-3mm}
\label{tab:ablation3}
\small
\begin{tabular}{@{}cc@{}}
\toprule
Number of stage & ACC@1  \\ \midrule
0               &65.33  \\
1              &66.03                   \\
2              &66.94                   \\
3              &67.08                   \\
4              &\textbf{67.13}                   \\ \bottomrule
\end{tabular}
\end{minipage}\hfill
\begin{minipage}[t]{0.32\linewidth}
\centering
\captionof{table}{Effect of CWRD (Class-Wise Relation Decoupling) and SWRD (Sample-Wise Relation Decoupling) in DFRA (Decoupled Finegrained Relation Alignment).}
\vspace{-3.4pt}
\label{tab:ablation1}
\footnotesize
\begin{tabular}{@{}cccc@{}}
\toprule
MSDF Module & CWRD
& SWRD
& Acc@l \\ \midrule
\checkmark           & $\times$                                       & $\times$                                        & 66.76 \\
\checkmark           & $\times$                                       & \checkmark                                         & 66.98 \\
\checkmark           & \checkmark                                        & $\times$                                        & 66.93 \\
\checkmark           & \checkmark                                        & \checkmark                                         & \textbf{67.13} \\ \bottomrule
\end{tabular}
\end{minipage}\hfill
\begin{minipage}[t]{0.42\linewidth}
\centering
\captionof{table}{More ablation studies in feature and logit levels. 
}
\vspace{-6.4pt}
\label{tab:ablation2}
\footnotesize
\setlength{\tabcolsep}{7.4pt}
\renewcommand{\arraystretch}{0.73}
\begin{tabular}{@{}ccccc@{}}
\toprule
Feature level & Logit level & MSDF & DFRA & Acc@1 \\ \midrule
\checkmark              &$\times$             &$\times$      &$\times$      &65.98       \\
\checkmark              &$\times$             &\checkmark      &$\times$      &66.43       \\
\checkmark              &$\times$             &$\times$      &\checkmark      &66.60       \\
\checkmark              &$\times$             &\checkmark      &\checkmark      &66.96       \\
$\times$              &\checkmark             &$\times$      &$\times$      &65.23       \\
$\times$              &\checkmark             &$\times$      &\checkmark      &65.33       \\
\checkmark              &\checkmark             &$\times$      &$\times$      &66.68       \\
\checkmark              &\checkmark             &\checkmark      &$\times$      &66.76       \\
\checkmark              &\checkmark             &$\times$      &\checkmark      &66.91       \\
\checkmark              &\checkmark             &\checkmark      &\checkmark      &\textbf{67.13}      \\   \bottomrule
\end{tabular}
\end{minipage}
\vspace{-3mm}
\end{table*}

\subsection{Results and Analysis}
\textbf{Results on CIFAR-100} \ We first conduct experiments on the CIFAR-100 dataset. Comparisons with the baselines are presented in Table~\ref{tab:cifar100}. It can be observed that our method can improve the performance of commonly used CNNs-based student models by 0.33\% to 1.43\%. Moreover, our method achieves remarkable results on Transformers-based student models, especially on the Mixer-B/16 and Deit-t pair, where the accuracy is raised by 4.86\%. Compared with KD, DKD, RKD and OFAKD, our method's improvement ranges from 0.09\% to 4.86\%. For the less prevalent MLPs-based student models, our method also achieves an improvement in accuracy by 0.20\% to 0.93\% compared with OFAKD. Moreover, the scarcely explored Mambas-based student models are significantly improved by our method, further verifying its effectiveness and generality. Overall, our proposed method achieves state-of-the-art performance on all different student architectures.


\begin{figure}[t]
  \centering
  \includegraphics[width=\linewidth]{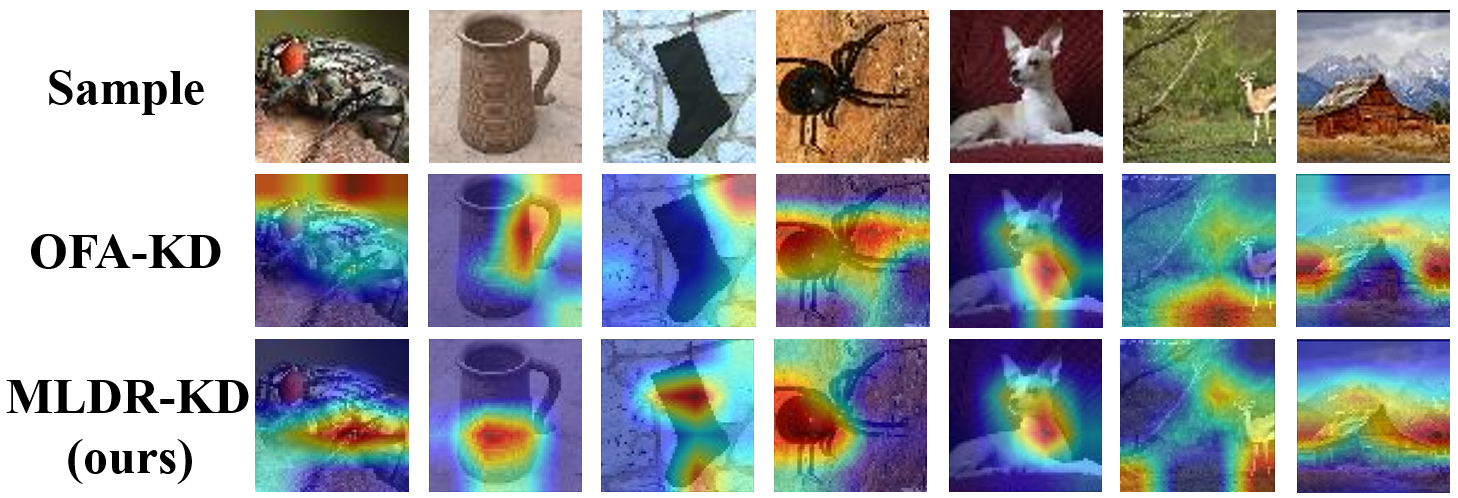}
  \vspace{-6mm}
  \caption{Comparisons of feature visualizations between OFA-KD and our MLDR-KD. The teacher is Vision Mamba Tiny, the student is ResNet-18. Clearly, our approach makes the student model more focused on the target across various samples.}
  \label{fig:featuremap2}
  \vspace{-4mm}
\end{figure}

\textbf{Results on Tiny-ImageNet} \ To assess the capability of our approach in coping with larger datasets, we expand experiments to the Tiny-ImageNet dataset. To align with the CIFAR-100 dataset, we select corresponding teacher and student models from CNNs-based, Transformers-based, MLPs-based, and Mambas-based models. As no previous baseline results are available, we compare our method with the baselines KD and OFAKD by reproducing them on Tiny-Imagenet. The results are showed in Table~\ref{tab:tinyimg}.

Our method exhibits more stable accuracy improvements than the baselines from the results on the Tiny-ImageNet dataset. Specifically, the accuracy improvement ranges from 0.26\% to 1.57\% on the CNNs-based student models and 0.20\% to 2.78\% on the Transformers-based student models, especially achieving an improvement of 2.78\% in the architecture pair ViM-T-to-Deit-T. Additionally, there is a significant improvement in our newly added student models T2t ViT-7 compared to the baseline methods. On the MLPs-based student models, our method could achieve an accuracy improvement of 0.12\% to 1.61\% compared to the baseline methods, particularly attaining the highest improvement of 1.61\% on the ResNet50 teacher model. For the latest Mamba-based student models, our method still presents considerable accuracy improvements.

Compared with the results on the CIFAR-100 dataset, the accuracy on the Tiny-ImageNet dataset exhibits advanced stability, indicating the advantages of our fine-grained design over traditional methods when dealing with larger datasets. Moreover, it can be observed that our method has more obvious improvements when applied to larger and more complicated models (such as ConvNeXt-T and ResNet50), validating that our method is potentially practical in further boosting off-the-shelf high-performance models, which are generally large and complicated. Similar to the case on the CIFAR-100 dataset, our method is applicable to the Mamba-based student models, further illustrating the generalization ability of the proposed heterogeneous distillation method.

\textbf{Visualization} \ A visual comparison between our method and baselines on the Tiny-Imagenet dataset is illustrated in Fig.~\ref{fig:featuremap2}. The students trained via our MLDR-KD can better learn from heterogeneous teachers. For example, even though the target occupies a small portion of the sample image, our student model is always able to focus on the information related to the target. The student model's attention does not diverge to distracting information. This is a strong indication that the student model, after our MLDR-KD, is well able to assimilate knowledge from heterogeneous teachers.



\subsection{Ablation study}

Ablative experiments are designed to verify the effectiveness of the proposed MLDR-KD, shown in Table.~\ref{tab:ablation3}, Table.~\ref{tab:ablation1} and Table.~\ref{tab:ablation2}. In this part, all
experiments are conducted on Tiny-Imagenet, with ViT-S as the
teacher model, Resnet18 as the student model.

\textbf{Number of stages of student} \ In Table.~\ref{tab:ablation3}, we conduct experiments to explore the impact of stages in student model. In order to accommodate different architectures, we divide the student model into a maximum of 4 stages. We chose stages 0 to 4 to compare the difference. We find that as the number of stages increases, the improvement effect of our methodology enhances. Four stages are the most potent.

\textbf{Validity of CWRD and SWRD in DFRA.} \ In order to verify the validity of proposed DFRA, we ablate our method in the presence of both feature and logit levels, shown in Table.~\ref{tab:ablation1}. From the results, CWRD and SWRD have improved by 0.22\% and 0.17\% relative to the original, respectively. Both of them can enhance 0.37\% in performance. Evidently, CWRD and SWRD in DFRA have an indispensable role to play. 

\textbf{Effect of our MLDR-KD in feature or logit level} \  Our MLDR-KD improves heterogeneous distillation in both feture and logit
levels. We study this improvement in this part. In Table.~\ref{tab:ablation2}, We ablate different modules at different levels. In a side-by-side comparison (\textit{e.g.} line 1-4), both of our methods MSDF and DFRA are effective when applied to only one level or to both levels. Vertical comparisons (\textit{e.g.} line 4 and 10) show that applying our methodology to multiple levels is the most effective.

%% file: sec/5_conclution.tex
\section{Conclusion}
\label{sec:conclusion}

In this paper, we propose Multi-Level Decoupled Relational Knowledge Distillation (MLDR-KD), a novel approach to balance the trade-off between dark knowledge and the confidence in the correct category of the teacher model for heterogeneous architectures. Specifically, DFRA is designed to align finegrained relationship for heterogeneous architectures in feature and logit level. The MSDF module is further introduced to improve DFRA performance by fusing feature maps of student in feature level. Extensive experiments show the robustness and generality of our MLDR-KD. Our future work involves how to efficiently take full advantage of feature information   
to further enhance the proposed MLDR-KD.